\pdfoutput=1

\documentclass[11pt]{article}

\usepackage{ACL2023}

\usepackage{times}
\usepackage{latexsym}

\usepackage[T1]{fontenc}

\usepackage[utf8]{inputenc}

\usepackage{microtype}

\usepackage{inconsolata}
\usepackage{multirow}
\usepackage{amsmath}
\usepackage{amssymb}
\usepackage{booktabs}
\usepackage{graphicx}
\usepackage{bbm}

\usepackage{diagbox}
\usepackage{makecell}

\title{CMOT: Cross-modal Mixup via Optimal Transport for Speech Translation}

\author{
    Yan Zhou$^{1}$,
    Qingkai Fang$^{1,2}$,
    Yang Feng$^{1,2}$\thanks{ $\;\;$Corresponding author: Yang Feng.} \\
    \textsuperscript{\rm1}Key Laboratory of Intelligent Information Processing \\ Institute of Computing Technology, Chinese Academy of Sciences (ICT/CAS) \\
    \textsuperscript{\rm2}University of Chinese Academy of Sciences, Beijing, China \\
    \texttt{2001zhouyan@gmail.com,} \quad \texttt{\{fangqingkai21b, fengyang\}@ict.ac.cn} \\
}

\begin{document}
\maketitle

\begin{abstract}
End-to-end speech translation (ST) is the task of translating speech signals in the source language into text in the target language. As a cross-modal task, end-to-end ST is difficult to train with limited data. Existing methods often try to transfer knowledge from machine translation (MT), but their performances are restricted by the modality gap between speech and text. In this paper, we propose \textbf{C}ross-modal \textbf{M}ixup via \textbf{O}ptimal \textbf{T}ransport (\textbf{CMOT}) to overcome the modality gap. We find the alignment between speech and text sequences via optimal transport and then mix up the sequences from different modalities at a token level using the alignment. Experiments on the MuST-C ST benchmark demonstrate that CMOT achieves an average BLEU of 30.0 in 8 translation directions, outperforming previous methods. Further analysis shows CMOT can adaptively find the alignment between modalities, which helps alleviate the modality gap between speech and text. \footnote{Code is publicly available at \url{https://github.com/ictnlp/CMOT}.}
\end{abstract}


\section{Introduction}
Speech translation (ST) is the task to translate speech signals in one language into text in another. Conventional ST systems work in a cascaded mode \citep{stentiford1988machine, waibel1991janus}, by combining automatic speech recognition (ASR) and machine translation (MT), which may suffer from error propagation and high latency. To solve these problems, end-to-end ST is proposed \citep{berard2016listen, duong2016attentional}. End-to-end ST directly generates the target text directly from the source audio without the intermediate transcript, which has drawn more attention in recent years \citep{vila2018end, salesky2019fluent, di2019adapting, di2019enhancing, inaguma2020espnet,wang2020fairseq, zhao2021neurst, dinh2022tackling, duquenne2022t}.
\par

End-to-end ST has to perform cross-lingual and cross-modal transformation simultaneously, hence bringing more challenges than ASR or MT. Unfortunately, it is harder to collect parallel speech-to-text pairs than text-to-text pairs, which makes ST suffer from the under-fitting problem. To solve the data shortage problem, some researchers propose methods to introduce text-to-text parallel data of MT into ST, including pretraining \citep{alinejad2020effectively, zheng2021fused, xu2021stacked}, multi-task learning \citep{le2020dual, vydana2021jointly, tang2021general}, knowledge distillation \citep{liu19d_interspeech, gaido2020end, inaguma2021source}, etc. In this way, the knowledge of MT can be transferred to ST, helping learn a better cross-lingual transformation. 

Although these methods of transfer learning have shown improvements in the translation quality, ST is joint learning of cross-lingual and cross-modal transformation, and the effectiveness of transfer learning is also based on the assumption that the cross-modal source will be projected to the common representation space. Therefore, the mixup strategy is employed to further improve ST by reducing the gap between the speech and text representation spaces. Previous methods with mixup strategy usually perform word-level mixup between aligned tokens in different modalities \citep{fang-etal-2022-stemm}, hence needing the help of external alignment tools. However, the alignment tools are not always available and require large amounts of additional annotated data for training, so how to obtain such alignments remains a problem for most languages.

\par


\par
Illuminated by the application of the optimal transport theory in finding the alignment between languages or modalities in recent years \citep{chen2020uniter, gu2022improving}, we propose \textbf{C}ross-modal \textbf{M}ixup via \textbf{O}ptimal \textbf{Tr}ansport \textbf{(CMOT)}. We use optimal transport to adaptively find the alignment between speech and text, and use this alignment to achieve token-level mixup. The mixup sequence will serve as a medium between the speech sequence and text sequence to realize cross-modal knowledge transfer. Experiments on the MuST-C dataset show that CMOT achieves an average BLEU of 30.0. In addition, we prove that optimal transport can help find cross-modal alignments and CMOT can help alleviate the modality gap in ST.

\section{Method}
\begin{figure}[t]
    \centering
    \includegraphics[width=\linewidth]{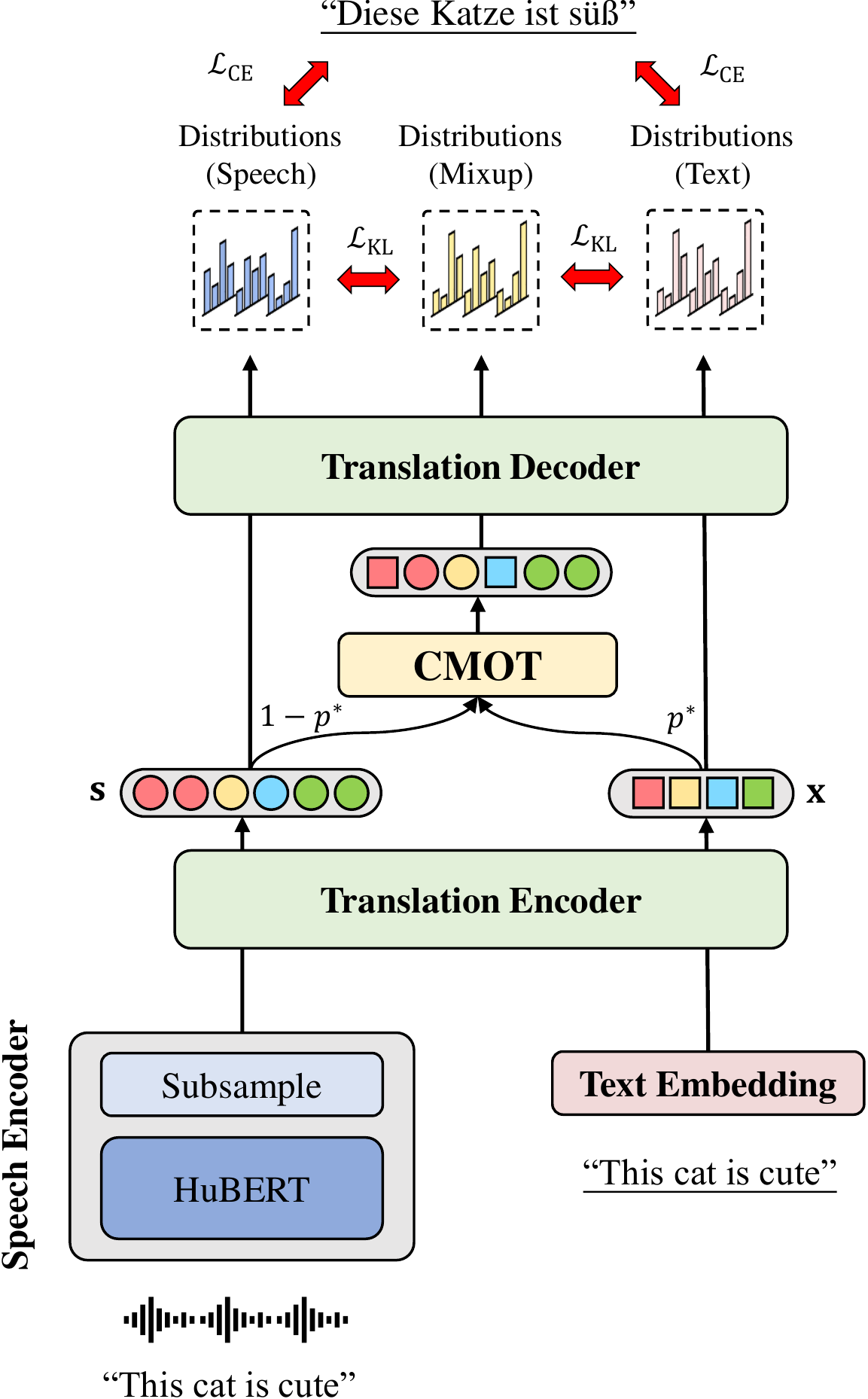}
    \caption{Overview of our proposed architecture with CMOT. }
    \label{fig:method}
\end{figure}

In this section, we will describe our proposed \textbf{C}ross-modal \textbf{M}ixup via \textbf{O}ptimal \textbf{Tr}ansport \textbf{(CMOT)}.
We find the alignment between speech and text sequences via OT, and mix up the two unimodal sequences to get a cross-modal mixed sequence. We predict the translation with all these sequences and regularize their outputs. Figure \ref{fig:method} illustrates the overview of our proposed method.


\subsection{Problem Formulation}
\label{sec:problem-formulation}
Generally, a speech translation corpus contains triplets of  source speech $\mathbf{s}$, its transcript $\mathbf{x}$ and translation $\mathbf{y}$, which can be denoted as $\mathcal{D}=\{(\mathbf{s},\mathbf{x},\mathbf{y})\}$. Given the corpus, an end-to-end ST system directly converts speech signals $\mathbf{s}$ into text translation $\mathbf{y}$ without generating the intermediate transcript $\mathbf{x}$.

\subsection{Model Architecture}
\label{sec:model-architecture}
Our model consists of four modules: the speech encoder, the text embedding layer, the translation encoder and the translation decoder. 
We first get parallel speech and text sequences with the speech encoder and the text embedding layer, and feed them into the translation encoder separately. Then we mix up the two encoder outputs with CMOT, and feed both the unimodal sequences and cross-modal mixed sequence to predict the translation.

\noindent\textbf{Speech Encoder}~
The speech encoder extracts the low-level features from the raw speech input. We use HuBERT \citep{hsu2021hubert} model stacked with a sub-sampler as the speech encoder. The HuBERT model is a self-supervised pretrained speech model and can improve the performance of ST systems as shown in recent works \citep{zhang2022improving, feng2022superb}. It consists of a convolutional feature extractor and a BERT \citep{devlin2019bert} style encoder. The sub-sampler consists of two convolution layers and is designed to reduce the length of the speech sequence.

\noindent\textbf{Text Embedding}~
The text embedding layer embeds tokenized text into a sequence of embeddings.

\noindent\textbf{Translation Encoder}~
The joint translation encoder receives either speech or text embeddings and learns further semantic information. It is composed of $N_e$ transformer \citep{vaswani2017attention} encoder layers.

\noindent\textbf{Translation Decoder}~
The joint translation decoder receives either unimodal sequences of speech and text or the cross-modal mixed sequence and predicts the translation $\mathbf{y}$. It is composed of $N_d$ transformer decoder layers.


\subsection{Cross-modal Mixup via Optimal Transport (CMOT)}
\label{sec:cmot}
We mix up the speech sequence and text sequence with CMOT. For a speech sequence $\mathbf{H}^s$ and the corresponding text sequence $\mathbf{H}^x$, we first get the alignment $\mathrm{A}$ between them via optimal transport (OT), and then mix up $\mathbf{H}^s$ and $\mathbf{H}^x$ with the help of $\mathrm{A}$ to get a mixed sequence $\mathbf{H}'$.

\noindent\textbf{Optimal Transport}~
The Optimal Transport (OT) problem is a classical mathematical problem and is nowadays used as a geometric tool to compare probability distributions \citep{peyre2019computational}.
It focuses on the scheme with the minimum cost of transferring one distribution to another under a given transfer cost.
Consider the following discrete transport problem: for two independent probability distributions $P$ and $Q$,
\begin{equation}
    \begin{aligned}
    &P = \{(w_i, m_i)\}_{i=1}^{n}, \quad s.t. \sum_{i}m_i=1; \\
    &Q = \{(\hat{w_j}, \hat{m_j})\}_{j=1}^{\hat{n}}, \quad s.t. \sum_{j}\hat{m_j}=1,
    \end{aligned}
\end{equation}
where each point $w_i \in \mathbb{R}^d$ has a weight $m_i \in [0, \infty)$. Given a transfer cost function $c(w_i, \hat{w_j})$ and denoting the mass transported from $w_i$ to $\hat{w_j}$ by $\mathbf{T}_{ij}$, the transport cost can be defined as:
\begin{equation}
    \begin{aligned}
        \mathcal{D}(P, Q) &= \min_{\mathbf{T} \geq 0} \sum_{i, j}\mathbf{T}_{ij}c(w_i, \hat{w_j}) \\
        s.t. \sum_{j=1}^{\hat{n}} \mathbf{T}_{ij} &= m_i, \forall i \in \{1, \dots, n \}, \\
        \sum_{i=1}^{n} \mathbf{T}_{ij} &= \hat{m_j}, \forall j \in \{1, \dots, \hat{n} \}.
    \end{aligned}
\end{equation}

\noindent\textbf{Relaxed OT with Window Strategy}~
\label{window}
If we regard the speech sequence and the text sequence as two independent distributions, we can use OT to measure the distance between them. For a speech sequence $\mathbf{H}^s = (h_1^s, \dots, h_n^s)$, a text sequence $\mathbf{H}^x = (h_1^x, \dots, h_{\hat{n}}^x)$, and a cost function $c$, we define the transport problem as:
\begin{equation}
    \begin{aligned}
        \mathcal{D}(\mathbf{H}^s, \mathbf{H}^x) &= \min_{\mathbf{T} \geq 0} \sum_{i, j}\mathbf{T}_{ij}c(h_i^s, h_j^x) \\
        s.t. \sum_{j=1}^{\hat{n}} \mathbf{T}_{ij} &= m_i^s, \forall i \in \{1, \dots, n \}, \\
        \sum_{i=1}^{n} \mathbf{T}_{ij} &= m_j^x, \forall j \in \{1, \dots, \hat{n} \}.
    \end{aligned}
\end{equation}
Here we use the Euclidean distance as the cost function $c$, and norm as the mass $m^s$ and $m^x$ inspired by \citet{DBLP:journals/corr/SchakelW15} and \citet{yokoi2020word}'s discovery that important tokens have larger norms. The solutions to the OT problem include some accurate algorithms like Sinkhorn \citep{cuturi2013sinkhorn} algorithm and IPOT \citep{xie2020fast}, which will bring great time complexity. \citet{kusner2015word} proposed a relaxed moving distance that removes one of the two constraints to obtain a lower bound of the accurate solution. Following this work, we remove the second constraint, and our transport problem thus becomes:
\begin{equation}
    \begin{aligned}
        \mathcal{D}^*(\mathbf{H}^s, \mathbf{H}^x) &= \min_{\mathbf{T} \geq 0} \sum_{i,j} \mathbf{T}_{ij}c(h_i^s, h_j^x) \\
        s.t. \sum_{j=1}^{\hat{n}} \mathbf{T}_{ij} &= m_i^s, \forall i \in \{1, \dots, n \}.
    \end{aligned}
    \label{equa:ot}
\end{equation}
This relaxed problem yields a lower bound of the original problem. We find that this relaxed OT improves the training speed without degrading the performance, as shown in Section \ref{sec:effect_ot}. Now the optimal solution for each speech token $h_i^s$ is to move all its mass to the closest text token $h_j^x$, so the transportation matrix becomes:
\begin{equation}
    \mathbf{T}_{ij}^*=
    \begin{cases}
        m_i^s & \text{if}\ j=\arg\min_j c(h^s_i, h^x_j) \\
        0 & \text{otherwise}
    \end{cases},
\end{equation}
This transport matrix $\mathbf{T}^*$ implies the alignment between $\mathbf{H}^s$ and $\mathbf{H}^x$. We define the cross-modal alignment as $\mathrm{A} = (a_1, \dots, a_n)$, where
\begin{equation}
    \begin{aligned}
        & a_i = \mathop{\mathrm{argmax}}\limits_{1 \leq j \leq \hat{n}} \mathbf{T}_{ij}^*.
    \end{aligned}
\end{equation}
\begin{figure}[t]
    \centering
    \includegraphics[width=\linewidth]{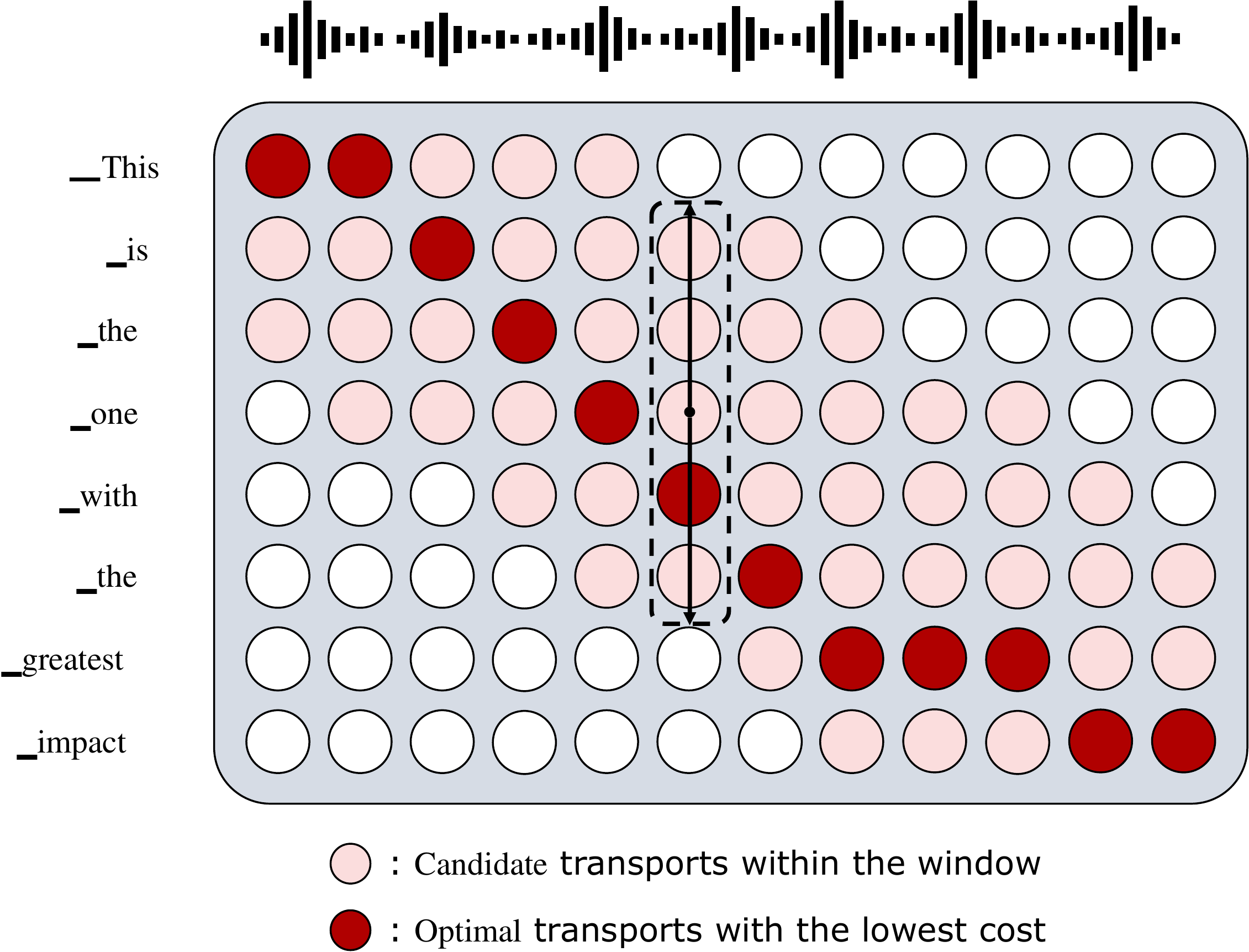}
    \caption{An example of OT from speech sequence to text sequence with window strategy. The x-axis is the speech sequence and the y-axis is the text sequence. White circles represent the transports that cannot be selected, pink ones represent the candidate transports and red ones represent the selected optimal transports. The dashed box represents the window, and we use $\mathrm{W}=2$ as an example here.}
    \label{fig:ot}
\end{figure}
Considering the monotonicity and locality of the alignment from the speech sequence to the text sequence, we apply a window strategy for this alignment. We constrain the index of the aligned target to the neighborhood of the diagonal. The window size is set as a constant $\mathrm{W}$ both to left and to right:
\begin{equation}
    \begin{aligned}
        &a_i = \mathop{\mathrm{argmax}}\limits_{\max{(1, \lambda i - \mathrm{W})} \ \leq \ j \ \leq \ \min{(\hat{n}, \lambda i + \mathrm{W})}} \mathbf{T}_{ij}^*, \\
        &\lambda = \hat{n} / n.
    \end{aligned}
\end{equation}
Figure \ref{fig:ot} shows how the relaxed OT with window strategy works. Through the above steps, we get the alignment $\mathrm{A}$ in an adaptive manner.

\noindent\textbf{Cross-modal Mixup via OT}~
With the help of the alignment $\mathrm{A}$, we will be able to mixup the speech and text sequences at the token level. For a mixup probability $p^*$, we generate the mixup sequence $\mathbf{H}'= (h_1', \dots, h_n')$ by:
\begin{equation}
    \begin{aligned}
        h_i' = 
        \begin{cases}
            h_i^s, \ \ p > p^*, \\
            h_{a_i}^x, \ p \leq p^*.
        \end{cases}
    \end{aligned}
    \label{equa:hidden}
\end{equation}
where $p$ is sampled from a uniform distribution $U(0,1)$. We feed $\mathbf{H}^s$, $\mathbf{H}^x$, $\mathbf{H}'$ into the joint transformer to predict the translation. Note that the positions of OT and mixup can be different. Our scheme is calculating OT before the translation encoder and performing mixup with the encoder outputs, which brings better performance in the experiment. See more details in Section \ref{abl:positions}.

\subsection{Training Strategy}
\label{sec:training-strategy}
We follow the paradigm of pretraining-finetuning to train our model.

\noindent\textbf{Pretraining}~
We pretrain the text embedding and the joint transformer with an MT task. The training objective during this phase is the cross entropy loss:
\begin{equation}
    \begin{aligned}
    \mathcal{L}_\mathrm{MT}=-\mathbb{E}_{\mathbf{x},\mathbf{y}} \log P(\mathbf{y}|\mathbf{x}).
    \end{aligned}
\end{equation}

\noindent\textbf{Multitask Finetuning}~
During finetuning, we train the model with multiple tasks. We add the cross entropy loss of ST and MT to the final loss:
\begin{equation}
    \begin{aligned}
        \mathcal{L}_\mathrm{ST}&=-\mathbb{E}_{(\mathbf{s},\mathbf{x},\mathbf{y}) \in \mathcal{D}} \log P(\mathbf{y}|\mathbf{s}), \\
        \mathcal{L}_\mathrm{MT}&=-\mathbb{E}_{(\mathbf{s},\mathbf{x},\mathbf{y}) \in \mathcal{D}} \log P(\mathbf{y}|\mathbf{x}).
    \end{aligned}
\end{equation}
Meanwhile, we regularize the predictions of mixup and speech sequences by minimizing the Kullback-Leibler (KL) Divergence between the token-level probability distributions of them, and we do the same for the predictions of mixup and text sequences. For a speech prediction distribution $S$, a text prediction distribution $T$ and mixup prediction distribution $M$, we have:
\begin{equation}
    \begin{aligned}
    & \mathcal{L}_{\mathrm{KL}}(M, S) = \frac{1}{2} (\mathbb{D}_{\mathrm{KL}}(M||S) + \mathbb{D}_{\mathrm{KL}}(S||M)), \\
    & \mathcal{L}_{\mathrm{KL}}(M, T) = \frac{1}{2} (\mathbb{D}_{\mathrm{KL}}(M||T) + \mathbb{D}_{\mathrm{KL}}(T||M)).
    \end{aligned}
\end{equation}
Hence the final training objective is:
\begin{equation}
    \label{equa:objective}
    \mathcal{L} = \mathcal{L}_{\mathrm{ST}} + \mathcal{L}_{\mathrm{MT}} + \lambda \mathcal{L}_{\mathrm{KL}}(M, S) + \lambda \mathcal{L}_{\mathrm{KL}}(M, T),
\end{equation}
where $\lambda$ is the weight to control $\mathcal{L}_{\mathrm{KL}}$.

\section{Experiments}
\begin{table*}[t]
\centering
\small
\resizebox{\textwidth}{!}{
\begin{tabular}{l|c|cccccccc|c}
\toprule
\multirow{2}{*}{\textbf{Models}} & \multirow{2}{*}{\makecell{\textbf{Speech} \\ \textbf{Pretraining}}} & \multicolumn{9}{c}{\textbf{BLEU}} \\
& & En-De & En-Fr & En-Ru & En-Es & En-It & En-Ro & En-Pt & En-Nl & Avg. \\ 
\midrule
\multicolumn{11}{c}{\textbf{Base setting} (\emph{w/o external MT data})} \\
\midrule
Fairseq ST \citep{wang2020fairseq} & \texttimes & 22.7 & 32.9 & 15.3 & 27.2 & 22.7 & 21.9 & 28.1 & 27.3 & 24.8  \\
XSTNet \citep{ye21_interspeech} & \checkmark &    25.5 & 36.0 & 16.9 & 29.6 & 25.5 & 25.1 & 31.3 & 30.0 & 27.5  \\
STEMM \citep{fang-etal-2022-stemm} & \checkmark &        25.6 & 36.1 & 17.1 & 30.3 & 25.6 & 24.3 & 31.0 & 30.1 & 27.5 \\
ConST \citep{ye-etal-2022-cross} & \checkmark &   25.7 & 36.8 & 17.3 & 30.4 & 26.3 & 24.8 & 32.0 & 30.6 & 28.0 \\
HuBERT-Transformer & \checkmark &                 25.4 & 36.5 & 16.8 & 30.5 & 26.0 & 24.2 & 31.7 & 29.9 & 27.6 \\
CMOT & \checkmark &                               \textbf{27.0**} & \textbf{37.3**} & \textbf{17.9**} & \textbf{31.1**} & \textbf{26.9**} & \textbf{25.3**} & \textbf{32.7**} & \textbf{31.2**} & \textbf{28.7} \\ 
\midrule
\multicolumn{11}{c}{\textbf{Expanded setting} (\emph{w/ external MT data})} \\
\midrule
Chimera \citep{han2021learning} & \checkmark &    27.1 & 35.6 & 17.4 & 30.6 & 25.0 & 24.0 & 30.2 & 29.2 & 27.4 \\
XSTNet \citep{ye21_interspeech} & \checkmark &    27.1 & 38.0 & 18.5 & 30.8 & 26.4 & 25.7 & 32.4 & 31.2 & 28.8  \\
STEMM \citep{fang-etal-2022-stemm} & \checkmark &        28.7 & 37.4 & 17.8 & 31.0 & 25.8 & 24.5 & 31.7 & 30.5 & 28.4 \\
ConST \citep{ye-etal-2022-cross} & \checkmark &   28.3 & 38.3 & 18.9 & 32.0 & 27.2 & 25.6 & 33.1 & 31.7 & 29.4 \\
HuBERT-Transformer & \checkmark &                 27.5 & 38.3 & 18.8 & 32.5 & 26.9 & 25.3 & 32.3 & 30.8 & 29.1 \\
CMOT & \checkmark &                               \textbf{29.0**} & \textbf{39.5**} & \textbf{19.2**} & \textbf{32.8*} & \textbf{27.5**} & \textbf{26.0**} & \textbf{33.5**} & \textbf{32.1**} & \textbf{30.0} \\ 
\bottomrule
\end{tabular}}
\caption{BLEU scores on MuST-C \texttt{tst-COMMON} set. ``Speech Pretraining'' denotes using pretrained speech models. *, ** indicate the improvements over HuBERT-Transformer are statistically significant with p < 0.05 and p < 0.01.}
\label{tab:main}
\end{table*}

We conduct experiments under two settings, the base and the expanded. The former only uses ST data, while the latter uses external MT data additionally. We will describe the datasets we use, our experimental setups and our main results in this section.

\subsection{Datasets}
\label{sec:dataset}

\noindent\textbf{ST Datasets}~
We conduct experiments on the MuST-C\footnote{Here we refer to v1.0. \url{https://ict.fbk.eu/must-c/}} \citep{di2019must} dataset, a widely used benchmark for ST. It consists of translations from English (En) to 8 languages: German (De), French (Fr), Russian (Ru), Spanish (Es), Italian (It), Romanian (Ro), Portuguese (Pt) and Dutch (Nl). For each direction, it comprises at least 385 hours of audio recorded from TED Talks. We use \texttt{dev} set for validation and \texttt{tst-COMMON} set for evaluation.

\noindent\textbf{External MT Datasets}~
We incorporate external MT data to pretrain the text embedding and the joint transformer. For directions of En-De, En-Fr, En-Ru, En-Es and En-Ro we use data from WMT \citep{bojar-etal-2016-findings}. For En-It, En-Pt and En-Nl we use data from OPUS100\footnote{\url{http://opus.nlpl.eu/opus-100.php}} \citep{zhang2020improving}.

Table \ref{tab:datasets} shows the detailed statistics of the MuST-C, WMT and OPUS100 datasets we use.
\begin{table}[t]
    \centering
    \resizebox{\linewidth}{!}{
    \begin{tabular}{c|cr|cr}
        \toprule
         & \multicolumn{2}{c|}{\textbf{ST (MuST-C)}} & \multicolumn{2}{c}{\textbf{MT}} \\
         \textbf{En$\rightarrow$} & hours & \#sents & name & \#sents \\
        \midrule
         \textbf{De} & 408 & 234K & WMT16 & 4.6M \\
         \textbf{Fr} & 492 & 280K & WMT14 & 40.8M \\
         \textbf{Ru} & 489 & 270K & WMT16 & 2.5M \\
         \textbf{Es} & 504 & 270K & WMT13 & 15.2M \\
         \textbf{It} & 465 & 258K & OPUS100 & 1.0M \\
         \textbf{Ro} & 432 & 240K & WMT16 & 0.6M \\
         \textbf{Pt} & 385 & 211K & OPUS100 & 1.0M \\
         \textbf{Nl} & 442 & 253K & OPUS100 & 1.0M \\
        \bottomrule
    \end{tabular}}
    \caption{Statistics of the datasets.}
    \label{tab:datasets}
\end{table}

\subsection{Experimental Setups}
\label{sec:setting}

\noindent\textbf{Model Configurations}~
For the speech encoder, we use HuBERT\footnote{\url{https://github.com/facebookresearch/fairseq/tree/main/examples/hubert}} \citep{hsu2021hubert} base model, which is pretrained on LibriSpeech \citep{panayotov2015librispeech} with no finetuning and is one of the state-of-the-art pretrained audio models. Following the usual practice for ST, we stack two convolution layers of kernel size 5, stride size 2, padding 2 and hidden size 1024 after the HuBERT model as a sub-sampler. The text embedding is an embedding layer of dimension 512. For the joint transformer, we use $N_e=6$ encoder layers and $N_d=6$ decoder layers with 512 hidden units, 2048 feed-forward hidden units and 8 attention heads. We use fairseq\footnote{\url{https://github.com/facebookresearch/fairseq/}} \citep{ott2019fairseq} for implementation.

\noindent\textbf{Pre-processing}~
For speech input, we use the 16-bit 16kHz mono-channel raw audio, and we filter out samples with frames greater than 480k or less than 1k to ensure training efficiency. For transcripts and translations, we tokenize them using a unigram SentencePiece\footnote{\url{https://github.com/google/sentencepiece}} \citep{kudo2018sentencepiece} model with a vocabulary size of 10k, which is shared by the source language and the target language.

\noindent\textbf{Training}~
Our training process follows the pretraining-finetuning paradigm. We pretrain the text embedding layer, the translation encoder and decoder with an MT task. For the base setting, we only use the transcript-translation pairs with a learning rate of 2e-3, and warm-up steps of 8k. For the expanded setting we first pretrain them on external MT data with a learning rate of 7e-4 and warm-up steps of 8k, and then pretrain them on transcript-translation pairs with a learning rate of 1e-4 and warm-up steps of 4k. For both settings, we allow at most 33k input tokens per batch. 
\par
We set the learning rate to 1e-4 and warm-up steps to 10k to finetune the whole model. We train the entire model for 60k updates with a batch of 16M audio frames. We use an Adam optimizer \citep{adam} with $\beta_1=0.9, \beta_2=0.98$, dropout of 0.1 and the label smoothing value of 0.1. For CMOT, we set the weight $\lambda$ of KL loss to 2.0 (see more in Appendix \ref{sec:weight_kl}) and the mixup probability $p^*$ to 0.2 (see more in Section \ref{abl:mixup_prob}), and we use the window strategy for OT by default with window size $\mathrm{W}=10$. All models are trained on 4 Nvidia GeForce RTX 3090 GPUs.

\noindent\textbf{Inference}~
We average the checkpoints of the last 10 epochs for evaluation. We use beam search with a beam size of 8. We evaluate case-sensitive detokenized BLEU on MuST-C \texttt{tst-COMMON} set using sacreBLEU\footnote{\url{https://github.com/mjpost/sacrebleu}} \citep{post2018call}. We use sacreBLEU to test the significance of the results.

\noindent\textbf{Baselines}~ 
We compare our method with the Fairseq ST \citep{wang2020fairseq} baseline implemented with the same framework \citep{ott2019fairseq}.
We implement HuBERT-Transformer as a strong baseline, which has the same architecture as our method. The only difference is that HuBERT-Transformer takes speech as input and performs the ST task during fine-tuning. We train
this model for 40k steps, except for the En-Ru direction where we use a training step of 100k due to specific language complexities.

\par
We also compare our method with several robust end-to-end ST systems: Chimera \citep{han2021learning} which learns a shared semantic space for speech and text, XSTNet \citep{ye21_interspeech} which uses a progressive multi-task training strategy, STEMM \citep{fang-etal-2022-stemm} which applies word-level mixup strategy and ConST \cite{ye-etal-2022-cross} which applies contrastive learning strategy. These baselines all adopt the multi-task learning mode and have similar model architectures.

\subsection{Main Results}
\label{sec:result}
Table \ref{tab:main} shows the main results. HuBERT-Transformer, the baseline model we implement, achieves relatively high BLEU scores. For the base setting without external MT data, CMOT outperforms HuBERT-Transformer by 1.1 BLEU in average of 8 directions. For the expanded setting, CMOT outperforms HuBERT-Transformer by 0.9 BLEU in average. Besides, CMOT also outperforms other strong baselines.

\section{Analysis}
\subsection{Effect of OT}
\label{sec:effect_ot}
\begin{figure*}[t]
    \centering
    \includegraphics[width=\linewidth]{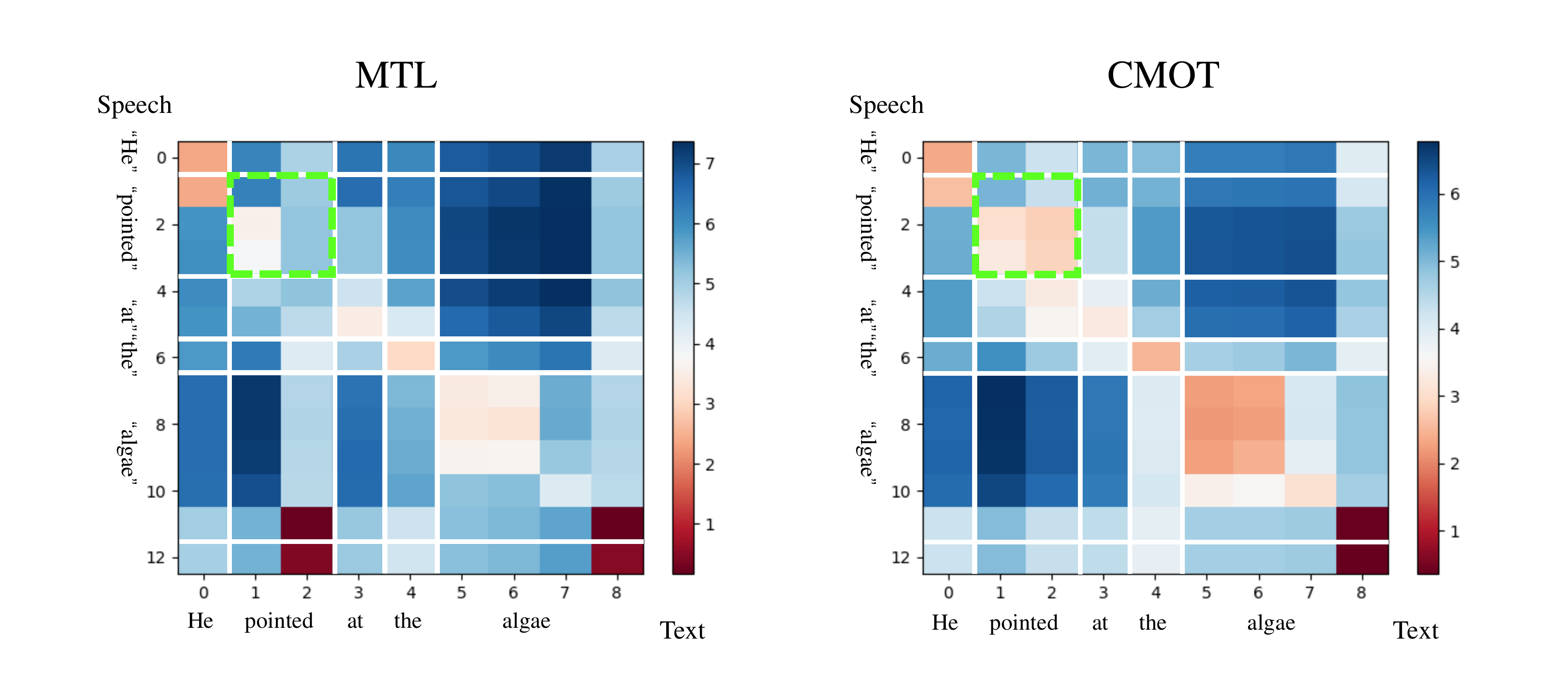}
    \caption{An example of how our CMOT can learn better alignment between speech and text. We draw the cost matrices in OT calculation for MTL and CMOT models. }
    \label{fig:matrix}
\end{figure*}


We use CMOT to learn the alignment between speech and text, and to improve the performance of the ST system. In this section, we compare CMOT with other models without OT or with other types of OT. All the models have the same architecture. MTL denotes the multi-task learning (ST and MT) model. Relaxed OT denotes the model trained with the same process of CMOT but without the window strategy in Section \ref{window}. IPOT denotes the model trained with the same process as CMOT but using IPOT \citep{xie2020fast} for OT calculation. The A-score is defined as the accuracy of our alignment $\mathrm{A} = (a_1, \dots, a_n)$ compared to the reference alignment $\mathrm{A^*} = (a^*_1, \dots, a^*_n)$:
\begin{equation}
    \text{A-score} = \sum_i \mathbbm{1}(a_i=a^*_i) / n,
\end{equation}
Here $A^*$ is obtained with Montreal Forced Aligner (MFA)\footnote{\url{https://montreal-forced-aligner.readthedocs.io/en/latest/}} \citep{mcauliffe2017montreal}. As shown in Table \ref{tab:ot_mfa}, we can see the result of relaxed OT is slightly lower than that of IPOT, but when the window strategy is used, its performance exceeds that of IPOT with a faster speed. CMOT achieves the best BLEU score, along with the best A-score of alignment, proving that CMOT not only takes advantage of this alignment but optimizes it as well.

\begin{table}[t]
    \centering
    \small
    \begin{tabular}{l|cc}
        \toprule
        \textbf{Model} & \textbf{BLEU$\uparrow$} & \textbf{A-score$\uparrow$} \\
        \midrule
        HuBERT-Transformer & 27.5 & 0.48 \\
        MTL                & 27.6 & 0.50 \\
        \midrule
        Relaxed OT         & 28.8 & 0.54 \\
        IPOT               & 28.9 & 0.55 \\
        \textbf{CMOT}      & \textbf{29.0} & \textbf{0.57} \\  
        \bottomrule
    \end{tabular}
    \caption{ST performances and alignment scores of different models on MuST-C En-De \texttt{tst-COMMON} set. These experiments are conducted under the expanded setting. }
    \label{tab:ot_mfa}
\end{table}

In addition, we observe the cost matrices of OT to verify that the cost matrix can be used for cross-modal alignment. We compare the cost matrices in MTL and CMOT, and analyze a typical example, as shown in Figure \ref{fig:matrix}. The redder grids indicate the distance between the speech token and the text token is smaller, while the bluer grids are the opposite. To better observe the word-level alignment, we use MFA to cut the sentences of speech and text into words, and the white solid lines denote the segmentation. For CMOT, we can see that the closer speech-text pairs basically lie on the diagonal, which is consistent with the alignment of the two sequences. But for MTL without the help of OT, the distance between speech and the corresponding word is not that close. Comparing the two green boxes in the figure, we can see CMOT has better alignment on the word ``pointed''.

\subsection{Effect of Cross-modal Mixup}
\label{sec:effect_miuxp}
\begin{figure}[t]
    \centering
    \includegraphics[width=\linewidth]{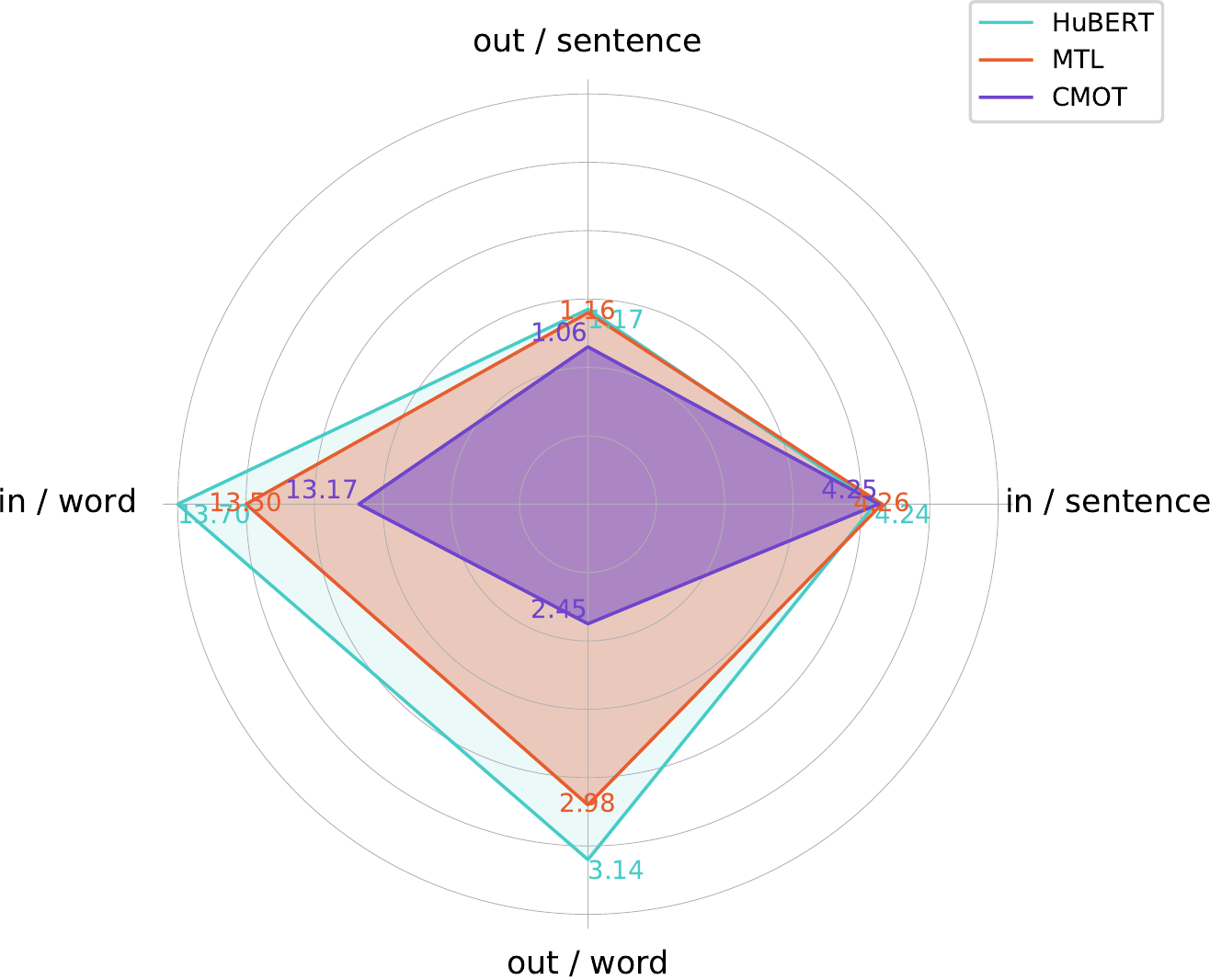}
    \caption{Modality gaps of different models. The data are calculated on the MuST-C En-De \texttt{dev} set with models of expanded setting. ``in'' / ``out'' denotes the encoder input / output sequences, while ``sentence'' / ``word'' denotes sentence-level / word-level Euclidean distance.}
    \label{fig:gap}
\end{figure}
To examine the effectiveness of the mixup strategy in bridging the modality gap, we compare the modality gap of CMOT and baseline models. We calculate the mean sentence-level distances and mean word-level distances of these models. We focus on the modality gap at the encoder input and output, and the results are shown in Figure \ref{fig:gap}. We can see that compared with the baseline model, CMOT has a smaller distance between modalities, which means our method effectively reduces the modality gap and improves the performance of ST.

\subsection{Ablation Study}
\noindent\textbf{Training Objectives}~
As a multi-task learning framework, our ST system's performance can be affected by the training objectives. Table \ref{tab:objective} shows the impact of training objectives on ST performance. Comparing Exp.I and II, We can empirically conclude that multi-task learning strategy alone helps little, while Exp.III suggests regularizing the output distributions is effective. Exp.IV, V, and VI show that calculating KL divergence for both mixup and speech outputs, and mixup and text outputs can bring the greatest improvement. In Exp.VII, we add the translation loss of the mixup sequence:
\begin{equation}
    \begin{aligned}
        &\mathcal{L}_\mathrm{Mixup}=-\mathbb{E}_{(\mathbf{s},\mathbf{x},\mathbf{y}) \in \mathcal{D}} \log P(\mathbf{y}|\mathbf{H}'),
    \end{aligned}
\end{equation}
where $\mathbf{H}'$ is the mixup sequence calculated as Equation (\ref{equa:hidden}). Comparing Exp.VI and VII, we can see $\mathcal{L}_\mathrm{Mixup}$ does not help. We also inspect the effect of OT loss which takes the value of $\mathcal{D}^*$ in Equation (\ref{equa:ot}) in Exp.VIII, and see a slight performance decline. Here OT is only used to help find the alignment, so OT loss is not indispensable during fine-tuning, and we conjecture adding OT loss may make it harder for the model to converge to good parameters, as the training objective is too complex.
\begin{table}[t]
    \centering
    \resizebox{\linewidth}{!}{
    \begin{tabular}{c|l|c}
        \toprule
        \textbf{\#Exp.} & \textbf{Training Objective} & \textbf{BLEU} \\
        \midrule
        \uppercase\expandafter{\romannumeral1} & $\mathcal{L}_\mathrm{ST}$ & 27.5 \\
        \uppercase\expandafter{\romannumeral2} & $\mathcal{L}_\mathrm{ST} + \mathcal{L}_\mathrm{MT}$ & 27.6 \\
        \uppercase\expandafter{\romannumeral3} & $\mathcal{L}_\mathrm{ST} + \mathcal{L}_\mathrm{MT} + \lambda\mathcal{L}_\mathrm{KL}(S, T)$ & 28.5 \\
        \uppercase\expandafter{\romannumeral4} & $\mathcal{L}_\mathrm{ST} + \mathcal{L}_\mathrm{MT} + \lambda\mathcal{L}_\mathrm{KL}(M, S)$ & 28.5 \\
        \uppercase\expandafter{\romannumeral5} & $\mathcal{L}_\mathrm{ST} + \mathcal{L}_\mathrm{MT} + \lambda\mathcal{L}_\mathrm{KL}(M, T)$ & 28.2 \\
        \uppercase\expandafter{\romannumeral6} & $\mathcal{L}_\mathrm{ST} + \mathcal{L}_\mathrm{MT} + \lambda\mathcal{L}_\mathrm{KL}(M, S) + \lambda\mathcal{L}_\mathrm{KL}(M, T)$ & \textbf{29.0} \\
        \uppercase\expandafter{\romannumeral7} & $\mathcal{L}_\mathrm{ST} + \mathcal{L}_\mathrm{MT} + \mathcal{L}_\mathrm{Mixup} + \lambda\mathcal{L}_\mathrm{KL}(M, S) + \lambda\mathcal{L}_\mathrm{KL}(M, T)$ & 28.7 \\
        \uppercase\expandafter{\romannumeral8} & $\mathcal{L}_\mathrm{ST} + \mathcal{L}_\mathrm{MT} + \lambda\mathcal{L}_\mathrm{KL}(M, S) + \lambda\mathcal{L}_\mathrm{KL}(M, T) + \mu\mathcal{L}_\mathrm{OT}$ & 28.6 \\
        \bottomrule
    \end{tabular}}
    \caption{ST performances with different training objectives on MuST-C En-De \texttt{tst-COMMON} set. These experiments are conducted under the expanded setting. $\mathcal{L}_\mathrm{Mixup}$ denotes the translation loss of the mixup sequence. We set the weights $\lambda=2$ and $\mu=0.1$.}
    \label{tab:objective}
\end{table}

\noindent\textbf{Positions of OT and Mixup}~
\label{abl:positions}
We investigate the impact of positions of OT and the mixup on our system. As shown in Table \ref{tab:position}, the best scheme is to calculate OT for encoder inputs and perform the mixup for encoder outputs. We believe there is more original alignment information at the lower layers of the model, which is better for OT calculation, while the hidden states of higher layers are more suitable for the mixup, as they have passed several layers of joint transformer and are closer in the representation space.
\begin{table}[t]
    \centering
    \small
    \begin{tabular}{l|cc}
        \toprule
        \diagbox{OT}{Mixup} & encoder in & encoder out \\
        \midrule
        encoder in  & 28.6 & \textbf{29.0} \\
        encoder out & 28.7 & 28.7 \\
        \bottomrule
    \end{tabular}
    \caption{BLEU scores of different positions of OT and mixup on MuST-C En-De \texttt{tst-COMMON} set. These experiments are conducted under the expanded setting.}
    \label{tab:position}
\end{table}

\begin{figure}[t]
    \centering
    \includegraphics[width=\linewidth]{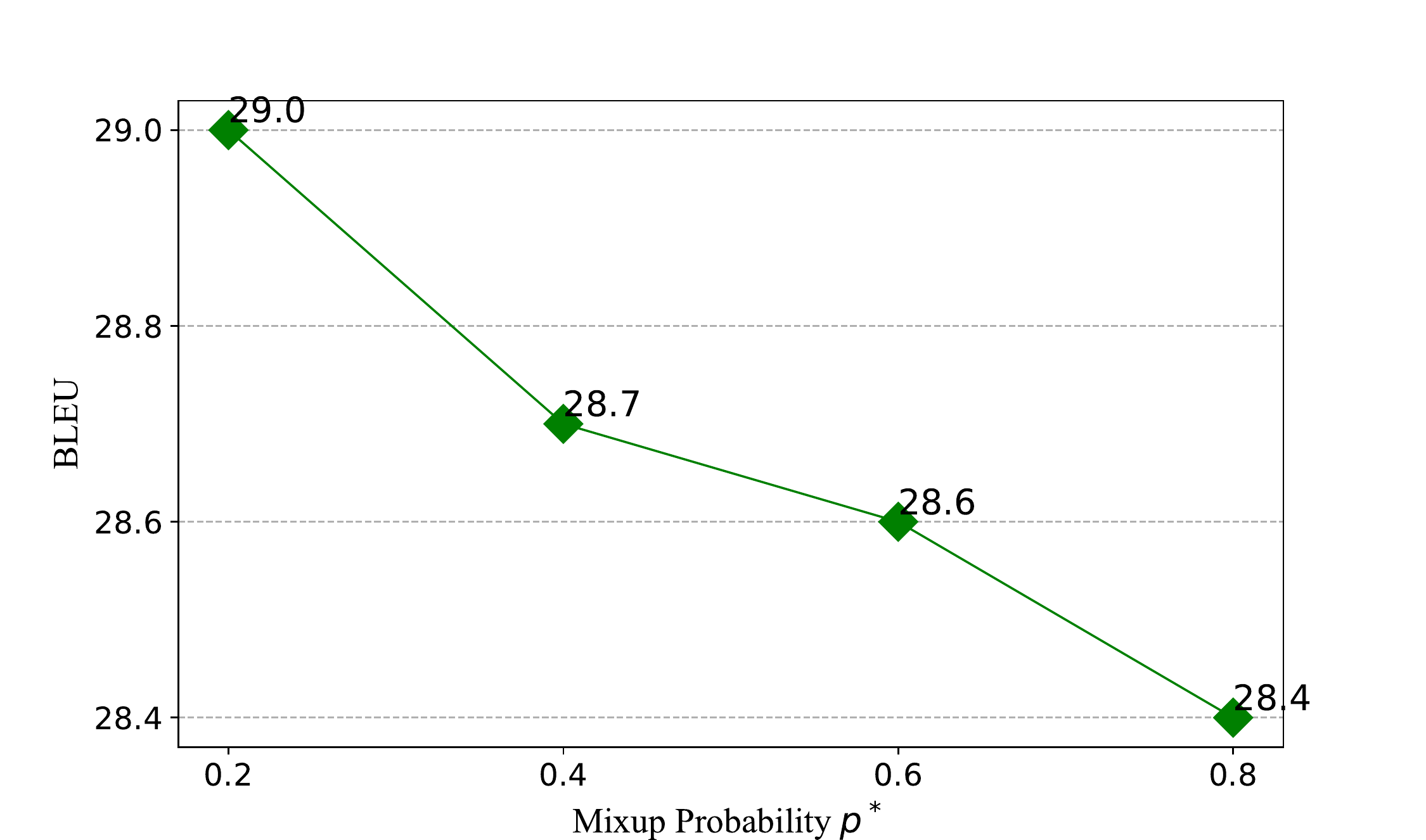}
    \caption{BLEU scores on MuST-C En-De \texttt{tst-COMMON} set with different mixup probability $p^*$. These experiments are conducted under the expanded setting. }
    \label{fig:prob}
\end{figure}

\noindent\textbf{Mixup Probability}~
\label{abl:mixup_prob}
We explore the impact of mixup probability on ST performance and the results are shown in Figure \ref{fig:prob}. Here $p^*$ denotes the proportion of text tokens in the mixup sequence. Previous works find a slightly larger probability is better, while we find $p^*=0.2$ improves ST performance the most. We think this is because CMOT needs to find the alignment between speech and text adaptively, so a larger proportion of text tokens in the mixup sequence might bring some noise.

\noindent\textbf{Speech Encoder}~
\label{abl:speech_encoder}
We investigate the impact of different speech encoders on our ST system. HuBERT \citep{hsu2021hubert} and wav2vec 2.0 \citep{baevski2020wav2vec} are both commonly used speech pretrained models. Table \ref{tab:speech_encoder} indicates that using HuBERT as a speech encoder can achieve higher performances than using wav2vec 2.0.
\begin{table}[t]
    \centering
    \resizebox{\linewidth}{!}{
    \begin{tabular}{l|cc}
        \toprule
        \textbf{Speech Encoder} & \textbf{Baseline} & \textbf{CMOT} \\
        \midrule
        wav2vec 2.0 \citep{baevski2020wav2vec} & 26.9 & 28.5 \\
        HuBERT \citep{hsu2021hubert}           & \textbf{27.5} & \textbf{29.0} \\
        \bottomrule
    \end{tabular}}
    \caption{BLEU scores with different speech encoders on MuST-C En-De \texttt{tst-COMMON} set. Here "baseline" represents the model with the same structure as CMOT but without mixup during training. These experiments are conducted under the expanded setting.}
    \label{tab:speech_encoder}
\end{table}
\par
In addition, in order to provide a fair comparison with other works using wav2vec 2.0, we conduct experiments on CMOT with wav2vec 2.0 as the speech encoder in more directions. As shown in Table \ref{tab:wav2vec2}, CMOT using wav2vec 2.0 still performs well, surpassing previous works in these three language pairs.
\begin{table}[t]
    \centering
    \small
    \resizebox{\linewidth}{!}{
    \begin{tabular}{l|ccc}
        \toprule
        \textbf{Model} & \textbf{En-De} & \textbf{En-Fr} & \textbf{En-Es} \\
        \midrule
        Chimera            & 27.1 & 35.6 & 30.6 \\
        XSTNet             & 27.1 & 38.0 & 30.8 \\
        STEMM              & 28.7 & 37.4 & 31.0 \\
        ConST              & 28.3 & 38.3 & 32.5 \\
        W2V2-Transformer   & 26.9 & 36.6 & 30.0 \\
        \textbf{CMOT-W2V2} & \textbf{28.5} & \textbf{39.1} & \textbf{32.6} \\
        \bottomrule
    \end{tabular}}
    \caption{BLEU scores with different models using wav2vec 2.0 on MuST-C \texttt{tst-COMMON} set. "CMOT-W2V2" refers to our CMOT model with wav2vec 2.0 as the speech encoder. "W2V2-Transformer" refers to the model with the same structure as CMOT-W2V2 but without mixup during training, and we directly report the results given in \citet{fang-etal-2022-stemm}. Here the experiments of W2V2-Transformer and CMOT-W2V2 are both conducted under the expanded setting.}
    \label{tab:wav2vec2}
\end{table}

\begin{figure}[t]
    \centering
    \includegraphics[width=\linewidth]{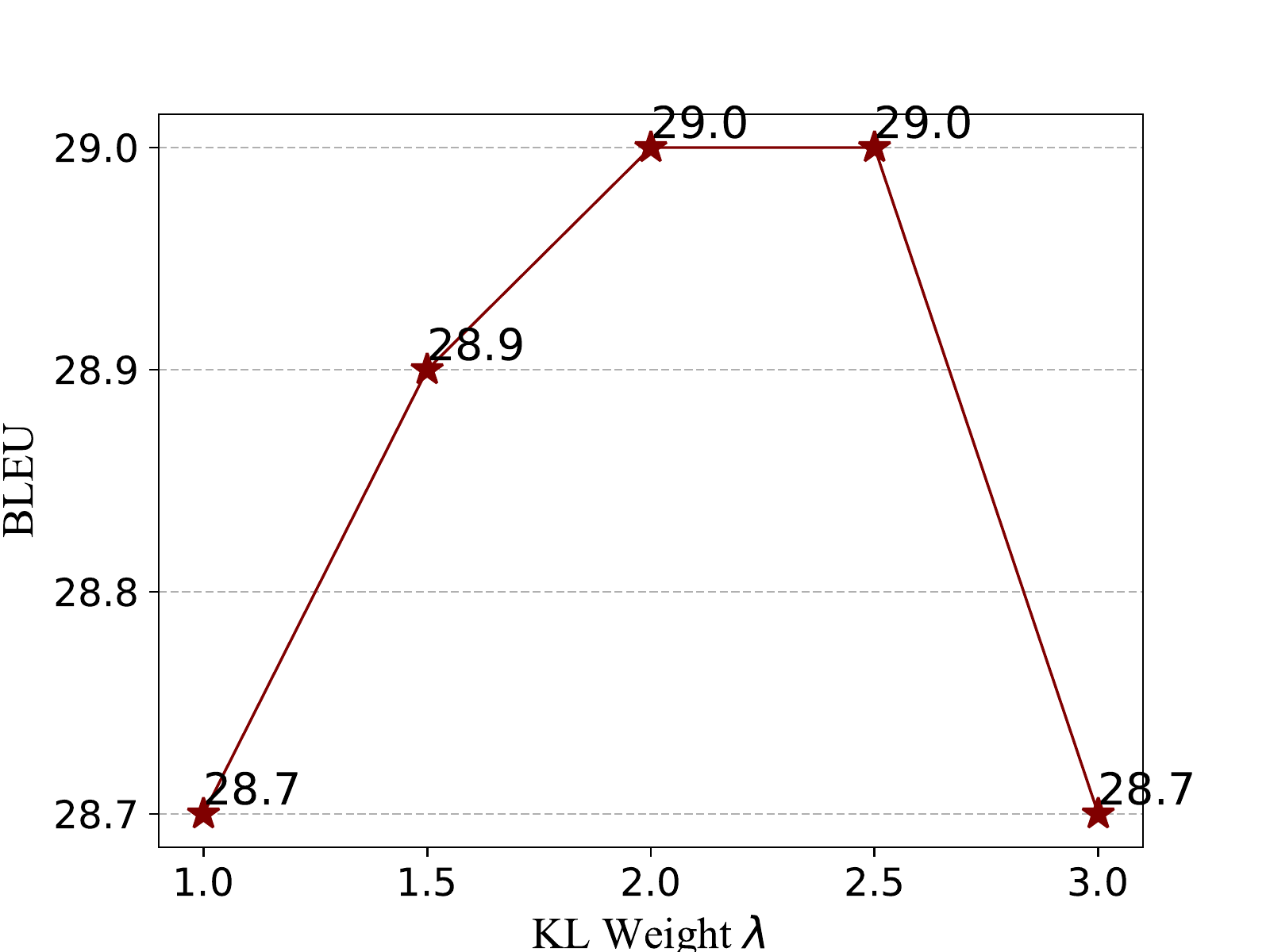}
    \caption{BLEU scores with different KL weight $\lambda$ on MuST-C En-De \texttt{tst-COMMON} set. Here the x-axis is the weight of KL loss. These experiments are all conducted under the expanded setting. We set the mixup probability $p^*=0.2$.}
    \label{fig:weight}
\end{figure}

\noindent\textbf{Weight of KL Loss}~
\label{sec:weight_kl}
Our finetuning process follows a multi-task learning approach, where the weight of each training objective can affect the overall performance. For the KL loss in Equation \ref{equa:objective}, we choose several different weight values $\lambda$ ranging from 1.0 to 3.0, and find when $\lambda=2.0$ or $\lambda=2.5$ the performance is best, as shown in Figure \ref{fig:weight}. When the weight of KL loss is too small, the regularization effect of the KL objective becomes little, while a large weight of KL loss can lead to performance degradation of the main tasks ST and MT. Hence, we suggest the weight be at a moderate value, and we choose $\lambda=2.0$ for our system.

\begin{figure}[t]
    \centering
    \includegraphics[width=\linewidth]{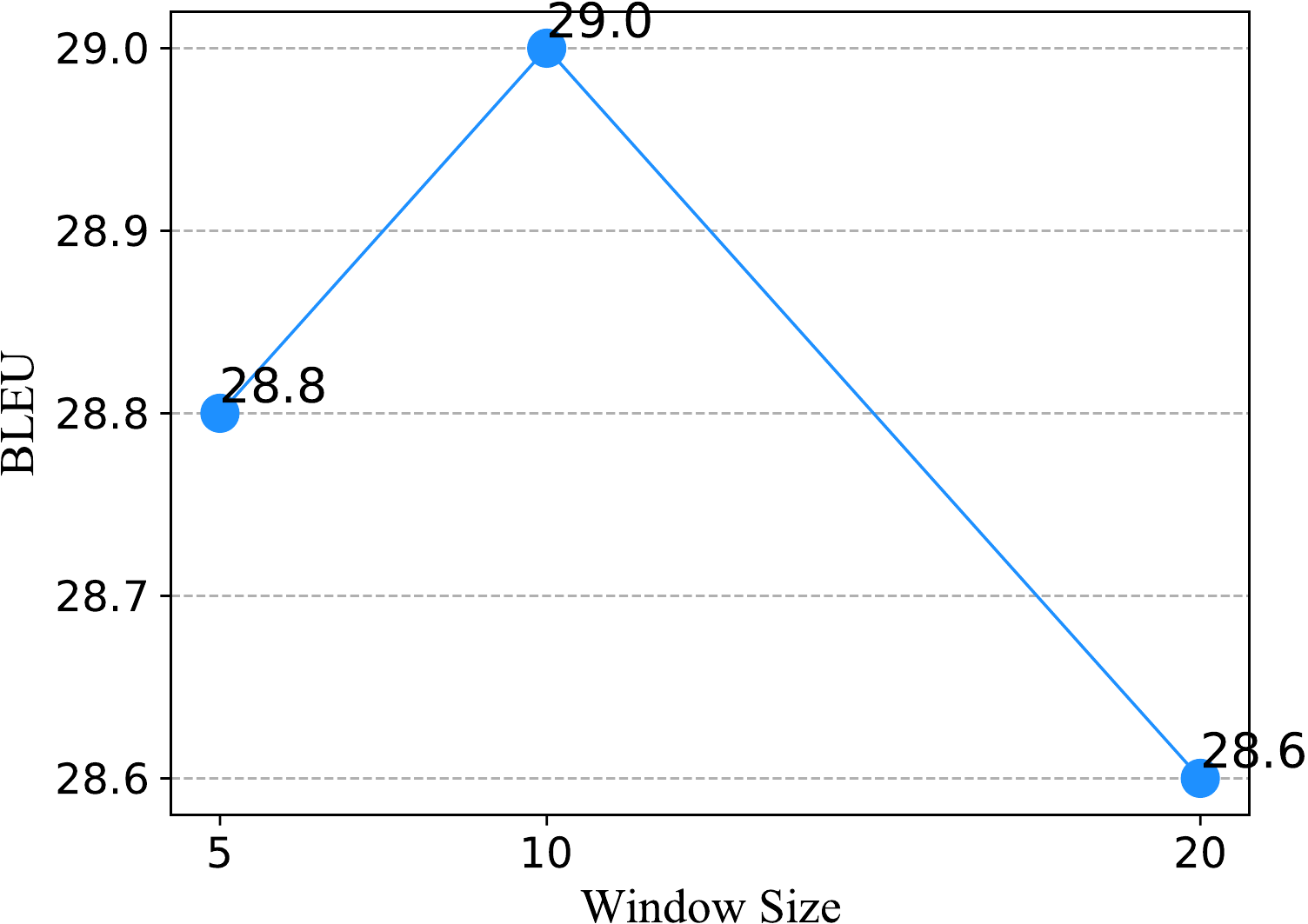}
    \caption{BLEU scores with different window sizes $\mathrm{W}$ on MuST-C En-De \texttt{tst-COMMON} set. These experiments are all conducted under the expanded setting.}
    \label{fig:window_size}
\end{figure}

\noindent\textbf{Window Size for OT}~
\label{sec:window_size}
We examine the impact of window size $\mathrm{W}$ on our ST system. As shown in Figure \ref{fig:window_size}, a small window size may restrict the model's ability to identify appropriate alignments, while a large window size may diminish its ability to capture local dependencies.

\section{Related Work}
\noindent\textbf{End-to-end ST}~
Traditional ST systems work in a cascaded mode \citep{stentiford1988machine, waibel1991janus}, with problems of error propagation and high latency. \citet{berard2016listen, duong2016attentional} proposed end-to-end ST without intermediate transcription, which has become the mainstream paradigm of ST in recent years \citep{vila2018end, salesky2019fluent, di2019adapting, di2019enhancing, inaguma2020espnet,wang2020fairseq, zhao2021neurst, dinh2022tackling, duquenne2022t}. An end-to-end ST system is difficult to train due to its cross-modal cross-lingual characteristics and the scarcity of parallel data, hence some techniques like pre-training \citep{bansal2019pre, stoian2020analyzing, wang2020bridging, alinejad2020effectively, zheng2021fused, xu2021stacked, zhang2022speechut}, self-training \citep{pino20_interspeech, wang21r_interspeech}, curriculum learning \citep{kano17_interspeech, wang2020curriculum}, multi-task learning \citep{le2020dual, vydana2021jointly, tang2021general}, data augmentation \citep{lam-etal-2022-sample, fang-and-feng-2023-back}, and knowledge distillation \citep{liu19d_interspeech, gaido2020end, inaguma2021source} are applied to improve its performance. To overcome the modality gap between speech and text, various methods are used. \citet{zhao2021mutual} proposed a mutual learning scenario. \citet{han2021learning} and \citet{duquenne:hal-03834732} projected speech and text into a common representation space. \citet{chen22r_interspeech} proposed the method of modality matching. \citet{ye-etal-2022-cross} applied the contrastive learning strategy. \citet{fang-and-feng-2023-understanding} proposed cross-modal regularization with scheduled sampling.

\noindent\textbf{Mixup}~
Our work draws on the strategy of mixup. Mixup is first proposed in \citet{zhang2018mixup} as a form of data augmentation to improve generalization and increase the robustness of neural networks by creating ``convex combinations of pairs of examples and their labels''. \citet{verma2019manifold} proposed the concept of manifold mixup, going a step further than surface-level mixup. In recent years, mixup is used to reduce the gap of representation spaces, such as cross-modal transfer \citep{so2022geodesic, ye2022dual} and cross-lingual transfer \citep{yang2021enhancing}. \citet{fang-etal-2022-stemm} proposed STEMM, applying mixup to end-to-end ST to overcome modality gap. Although both STEMM and our work use the mixup strategy, STEMM needs external alignment tools, while CMOT finds the alignment adaptively.

\noindent\textbf{Optimal Transport}~
Optimal transport is a classical mathematical problem. It is now commonly used to describe the transfer cost between two distributions. \citet{villani2009optimal} systematically demonstrates the theories and proofs of optimal transport. \citet{peyre2019computational} outlines the main theoretical insights that support the practical effectiveness of OT. There are some approximation algorithms for the OT problem, like the Sinkhorn algorithm \citep{cuturi2013sinkhorn} and IPOT \citep{xie2020fast} algorithm. \citet{kusner2015word} proposed a relaxed form of OT in order to measure the similarity between documents, which reduced the computational cost. \citet{chen2020uniter} used OT in image-text pre-training to ``explicitly encourage fine-grained alignment between words and image regions''. \citet{gu2022improving} applied the relaxed form of OT to machine translation and managed to ``bridge the gap between the semantic-equivalent representations of different languages''.

\section{Conclusion}
We propose \textbf{C}ross-modal \textbf{M}ixup via \textbf{O}ptimal \textbf{T}ransport (\textbf{CMOT}) to adaptively find the alignment between speech and text sequences, and to mix up the sequences of different modalities at the token level. Experiments on the MuST-C benchmark demonstrate the effectiveness of our method, which helps overcome the modality gap and thus improves the performance of end-to-end ST systems. We will continue to explore the application of this adaptive cross-modal alignment method in the future.

\section*{Limitations}


The CMOT method for speech translation is based on the idea of cross-modal knowledge transfer and the paradigm of multi-task learning, so it requires the transcripts of speech during training, which may not be applicable to some languages without transcribed text. Besides, this work mainly focuses on the finetuning of speech translation, and applications on the pretraining phase or on other tasks are in need of exploration in the future.

\section*{Ethics Statement}

Our proposed CMOT can help build a strong end-to-end ST system. It has prospects for application in many scenarios requiring speech translation, which helps people understand speech in a foreign language. Nevertheless, the results generated by end-to-end ST systems are not necessarily perfect, so people may not fully rely on the results for the time being.

\section*{Acknowledgements}
This work was supported by National Key R\&D Program of China(NO. 2018AAA0102502). We thank all the anonymous reviewers for their insightful and valuable comments.

\bibliography{anthology,custom}

\begin{thebibliography}{69}
\expandafter\ifx\csname natexlab\endcsname\relax\def\natexlab#1{#1}\fi

\bibitem[{Alinejad and Sarkar(2020)}]{alinejad2020effectively}
Ashkan Alinejad and Anoop Sarkar. 2020.
\newblock Effectively pretraining a speech translation decoder with machine
  translation data.
\newblock In \emph{Proceedings of the 2020 Conference on Empirical Methods in
  Natural Language Processing (EMNLP)}, pages 8014--8020.

\bibitem[{Baevski et~al.(2020)Baevski, Zhou, Mohamed, and
  Auli}]{baevski2020wav2vec}
Alexei Baevski, Yuhao Zhou, Abdelrahman Mohamed, and Michael Auli. 2020.
\newblock wav2vec 2.0: A framework for self-supervised learning of speech
  representations.
\newblock \emph{Advances in Neural Information Processing Systems},
  33:12449--12460.

\bibitem[{Bansal et~al.(2019)Bansal, Kamper, Livescu, Lopez, and
  Goldwater}]{bansal2019pre}
S~Bansal, H~Kamper, K~Livescu, A~Lopez, and S~Goldwater. 2019.
\newblock Pre-training on high-resource speech recognition improves
  low-resource speech-to-text translation.
\newblock \emph{NAACL 2019}.

\bibitem[{B{\'e}rard et~al.(2016)B{\'e}rard, Pietquin, Besacier, and
  Servan}]{berard2016listen}
Alexandre B{\'e}rard, Olivier Pietquin, Laurent Besacier, and Christophe
  Servan. 2016.
\newblock Listen and translate: A proof of concept for end-to-end
  speech-to-text translation.
\newblock In \emph{NIPS Workshop on end-to-end learning for speech and audio
  processing}.

\bibitem[{Bojar et~al.(2016)Bojar, Chatterjee, Federmann, Graham, Haddow, Huck,
  Jimeno~Yepes, Koehn, Logacheva, Monz, Negri, N{\'e}v{\'e}ol, Neves, Popel,
  Post, Rubino, Scarton, Specia, Turchi, Verspoor, and
  Zampieri}]{bojar-etal-2016-findings}
Ond{\v{r}}ej Bojar, Rajen Chatterjee, Christian Federmann, Yvette Graham, Barry
  Haddow, Matthias Huck, Antonio Jimeno~Yepes, Philipp Koehn, Varvara
  Logacheva, Christof Monz, Matteo Negri, Aur{\'e}lie N{\'e}v{\'e}ol, Mariana
  Neves, Martin Popel, Matt Post, Raphael Rubino, Carolina Scarton, Lucia
  Specia, Marco Turchi, Karin Verspoor, and Marcos Zampieri. 2016.
\newblock \href {https://doi.org/10.18653/v1/W16-2301} {Findings of the 2016
  conference on machine translation}.
\newblock In \emph{Proceedings of the First Conference on Machine Translation:
  Volume 2, Shared Task Papers}, pages 131--198, Berlin, Germany. Association
  for Computational Linguistics.

\bibitem[{Chen et~al.(2020)Chen, Li, Yu, El~Kholy, Ahmed, Gan, Cheng, and
  Liu}]{chen2020uniter}
Yen-Chun Chen, Linjie Li, Licheng Yu, Ahmed El~Kholy, Faisal Ahmed, Zhe Gan,
  Yu~Cheng, and Jingjing Liu. 2020.
\newblock Uniter: Universal image-text representation learning.
\newblock In \emph{Computer Vision--ECCV 2020: 16th European Conference,
  Glasgow, UK, August 23--28, 2020, Proceedings, Part XXX}, pages 104--120.

\bibitem[{Chen et~al.(2022)Chen, Zhang, Rosenberg, Ramabhadran, Moreno, Bapna,
  and Zen}]{chen22r_interspeech}
Zhehuai Chen, Yu~Zhang, Andrew Rosenberg, Bhuvana Ramabhadran, Pedro~J. Moreno,
  Ankur Bapna, and Heiga Zen. 2022.
\newblock \href {https://doi.org/10.21437/Interspeech.2022-10937} {{MAESTRO:
  Matched Speech Text Representations through Modality Matching}}.
\newblock In \emph{Proc. Interspeech 2022}, pages 4093--4097.

\bibitem[{Cuturi(2013)}]{cuturi2013sinkhorn}
Marco Cuturi. 2013.
\newblock Sinkhorn distances: Lightspeed computation of optimal transport.
\newblock \emph{Advances in neural information processing systems}, 26.

\bibitem[{Devlin et~al.(2019)Devlin, Chang, Lee, and
  Toutanova}]{devlin2019bert}
Jacob Devlin, Ming-Wei Chang, Kenton Lee, and Kristina Toutanova. 2019.
\newblock Bert: Pre-training of deep bidirectional transformers for language
  understanding.
\newblock In \emph{Proceedings of the 2019 Conference of the North American
  Chapter of the Association for Computational Linguistics: Human Language
  Technologies, Volume 1 (Long and Short Papers)}, pages 4171--4186.

\bibitem[{Di~Gangi et~al.(2019{\natexlab{a}})Di~Gangi, Cattoni, Bentivogli,
  Negri, and Turchi}]{di2019must}
Mattia~A Di~Gangi, Roldano Cattoni, Luisa Bentivogli, Matteo Negri, and Marco
  Turchi. 2019{\natexlab{a}}.
\newblock Must-c: a multilingual speech translation corpus.
\newblock In \emph{2019 Conference of the North American Chapter of the
  Association for Computational Linguistics: Human Language Technologies},
  pages 2012--2017. Association for Computational Linguistics.

\bibitem[{Di~Gangi et~al.(2019{\natexlab{b}})Di~Gangi, Negri, and
  Turchi}]{di2019adapting}
Mattia~A Di~Gangi, Matteo Negri, and Marco Turchi. 2019{\natexlab{b}}.
\newblock Adapting transformer to end-to-end spoken language translation.
\newblock In \emph{INTERSPEECH 2019}, pages 1133--1137. International Speech
  Communication Association (ISCA).

\bibitem[{Di~Gangi et~al.(2019{\natexlab{c}})Di~Gangi, Negri, Cattoni, Dessi,
  and Turchi}]{di2019enhancing}
Mattia~Antonino Di~Gangi, Matteo Negri, Roldano Cattoni, Roberto Dessi, and
  Marco Turchi. 2019{\natexlab{c}}.
\newblock Enhancing transformer for end-to-end speech-to-text translation.
\newblock In \emph{Proceedings of Machine Translation Summit XVII: Research
  Track}, pages 21--31.

\bibitem[{Dinh et~al.(2022)Dinh, Liu, and Niehues}]{dinh2022tackling}
Tu~Anh Dinh, Danni Liu, and Jan Niehues. 2022.
\newblock Tackling data scarcity in speech translation using zero-shot
  multilingual machine translation techniques.
\newblock In \emph{ICASSP 2022-2022 IEEE International Conference on Acoustics,
  Speech and Signal Processing (ICASSP)}, pages 6222--6226. IEEE.

\bibitem[{Duong et~al.(2016)Duong, Anastasopoulos, Chiang, Bird, and
  Cohn}]{duong2016attentional}
Long Duong, Antonios Anastasopoulos, David Chiang, Steven Bird, and Trevor
  Cohn. 2016.
\newblock An attentional model for speech translation without transcription.
\newblock In \emph{Proceedings of the 2016 Conference of the North American
  Chapter of the Association for Computational Linguistics: Human Language
  Technologies}, pages 949--959.

\bibitem[{Duquenne et~al.(2022{\natexlab{a}})Duquenne, Gong, Sagot, and
  Schwenk}]{duquenne2022t}
Paul-Ambroise Duquenne, Hongyu Gong, Beno{\^\i}t Sagot, and Holger Schwenk.
  2022{\natexlab{a}}.
\newblock T-modules: Translation modules for zero-shot cross-modal machine
  translation.
\newblock \emph{arXiv preprint arXiv:2205.12216}.

\bibitem[{Duquenne et~al.(2022{\natexlab{b}})Duquenne, Gong, Sagot, and
  Schwenk}]{duquenne:hal-03834732}
Paul-Ambroise Duquenne, Hongyu Gong, Beno{\^i}t Sagot, and Holger Schwenk.
  2022{\natexlab{b}}.
\newblock \href {https://hal.inria.fr/hal-03834732} {{T-Modules: Translation
  Modules for Zero-Shot Cross-Modal Machine Translation}}.
\newblock In \emph{{EMNLP 2022 - 2022 Conference on Empirical Methods in
  Natural Language Processing}}, Abu Dhabi, United Arab Emirates.

\bibitem[{Fang and Feng(2023{\natexlab{a}})}]{fang-and-feng-2023-back}
Qingkai Fang and Yang Feng. 2023{\natexlab{a}}.
\newblock Back translation for speech-to-text translation without transcripts.
\newblock In \emph{Proceedings of the 61st Annual Meeting of the Association
  for Computational Linguistics}.

\bibitem[{Fang and Feng(2023{\natexlab{b}})}]{fang-and-feng-2023-understanding}
Qingkai Fang and Yang Feng. 2023{\natexlab{b}}.
\newblock Understanding and bridging the modality gap for speech translation.
\newblock In \emph{Proceedings of the 61st Annual Meeting of the Association
  for Computational Linguistics}.

\bibitem[{Fang et~al.(2022)Fang, Ye, Li, Feng, and Wang}]{fang-etal-2022-stemm}
Qingkai Fang, Rong Ye, Lei Li, Yang Feng, and Mingxuan Wang. 2022.
\newblock \href {https://doi.org/10.18653/v1/2022.acl-long.486} {{STEMM}:
  Self-learning with speech-text manifold mixup for speech translation}.
\newblock In \emph{Proceedings of the 60th Annual Meeting of the Association
  for Computational Linguistics (Volume 1: Long Papers)}, pages 7050--7062,
  Dublin, Ireland. Association for Computational Linguistics.

\bibitem[{Feng et~al.(2022)Feng, Dong, Yeh, Yang, Lin, Shi, Chang, Huang, Wu,
  Chang et~al.}]{feng2022superb}
Tzu-hsun Feng, Annie Dong, Ching-Feng Yeh, Shu-wen Yang, Tzu-Quan Lin, Jiatong
  Shi, Kai-Wei Chang, Zili Huang, Haibin Wu, Xuankai Chang, et~al. 2022.
\newblock Superb@ slt 2022: Challenge on generalization and efficiency of
  self-supervised speech representation learning.
\newblock \emph{arXiv preprint arXiv:2210.08634}.

\bibitem[{Gaido et~al.(2020)Gaido, Di~Gangi, Negri, and Turchi}]{gaido2020end}
Marco Gaido, Mattia~A Di~Gangi, Matteo Negri, and Marco Turchi. 2020.
\newblock End-to-end speech-translation with knowledge distillation: Fbk@
  iwslt2020.
\newblock In \emph{Proceedings of the 17th International Conference on Spoken
  Language Translation}, pages 80--88.

\bibitem[{Gu and Feng(2022)}]{gu2022improving}
Shuhao Gu and Yang Feng. 2022.
\newblock Improving zero-shot multilingual translation with universal
  representations and cross-mappings.
\newblock \emph{arXiv preprint arXiv:2210.15851}.

\bibitem[{Han et~al.(2021)Han, Wang, Ji, and Li}]{han2021learning}
Chi Han, Mingxuan Wang, Heng Ji, and Lei Li. 2021.
\newblock Learning shared semantic space for speech-to-text translation.
\newblock In \emph{Findings of the Association for Computational Linguistics:
  ACL-IJCNLP 2021}, pages 2214--2225.

\bibitem[{Hsu et~al.(2021)Hsu, Bolte, Tsai, Lakhotia, Salakhutdinov, and
  Mohamed}]{hsu2021hubert}
Wei-Ning Hsu, Benjamin Bolte, Yao-Hung~Hubert Tsai, Kushal Lakhotia, Ruslan
  Salakhutdinov, and Abdelrahman Mohamed. 2021.
\newblock Hubert: Self-supervised speech representation learning by masked
  prediction of hidden units.
\newblock \emph{IEEE/ACM Transactions on Audio, Speech, and Language
  Processing}, 29:3451--3460.

\bibitem[{Inaguma et~al.(2021)Inaguma, Kawahara, and
  Watanabe}]{inaguma2021source}
Hirofumi Inaguma, Tatsuya Kawahara, and Shinji Watanabe. 2021.
\newblock Source and target bidirectional knowledge distillation for end-to-end
  speech translation.
\newblock In \emph{Proceedings of the 2021 Conference of the North American
  Chapter of the Association for Computational Linguistics: Human Language
  Technologies}, pages 1872--1881.

\bibitem[{Inaguma et~al.(2020)Inaguma, Kiyono, Duh, Karita, Yalta, Hayashi, and
  Watanabe}]{inaguma2020espnet}
Hirofumi Inaguma, Shun Kiyono, Kevin Duh, Shigeki Karita, Nelson Yalta, Tomoki
  Hayashi, and Shinji Watanabe. 2020.
\newblock Espnet-st: All-in-one speech translation toolkit.
\newblock In \emph{Proceedings of the 58th Annual Meeting of the Association
  for Computational Linguistics: System Demonstrations}, pages 302--311.

\bibitem[{Kano et~al.(2017)Kano, Sakti, and Nakamura}]{kano17_interspeech}
Takatomo Kano, Sakriani Sakti, and Satoshi Nakamura. 2017.
\newblock \href {https://doi.org/10.21437/Interspeech.2017-944}
  {{Structured-Based Curriculum Learning for End-to-End English-Japanese Speech
  Translation}}.
\newblock In \emph{Proc. Interspeech 2017}, pages 2630--2634.

\bibitem[{Kingma and Ba(2015)}]{adam}
Diederik~P. Kingma and Jimmy Ba. 2015.
\newblock \href {http://arxiv.org/abs/1412.6980} {Adam: A method for stochastic
  optimization}.
\newblock In \emph{ICLR (Poster)}.

\bibitem[{Kudo and Richardson(2018)}]{kudo2018sentencepiece}
Taku Kudo and John Richardson. 2018.
\newblock Sentencepiece: A simple and language independent subword tokenizer
  and detokenizer for neural text processing.
\newblock In \emph{Proceedings of the 2018 Conference on Empirical Methods in
  Natural Language Processing: System Demonstrations}, pages 66--71.

\bibitem[{Kusner et~al.(2015)Kusner, Sun, Kolkin, and
  Weinberger}]{kusner2015word}
Matt Kusner, Yu~Sun, Nicholas Kolkin, and Kilian Weinberger. 2015.
\newblock From word embeddings to document distances.
\newblock In \emph{International conference on machine learning}, pages
  957--966. PMLR.

\bibitem[{Lam et~al.(2022)Lam, Schamoni, and Riezler}]{lam-etal-2022-sample}
Tsz~Kin Lam, Shigehiko Schamoni, and Stefan Riezler. 2022.
\newblock \href {https://doi.org/10.18653/v1/2022.acl-short.27} {Sample,
  translate, recombine: Leveraging audio alignments for data augmentation in
  end-to-end speech translation}.
\newblock In \emph{Proceedings of the 60th Annual Meeting of the Association
  for Computational Linguistics (Volume 2: Short Papers)}, pages 245--254,
  Dublin, Ireland. Association for Computational Linguistics.

\bibitem[{Le et~al.(2020)Le, Pino, Wang, Gu, Schwab, and Besacier}]{le2020dual}
Hang Le, Juan Pino, Changhan Wang, Jiatao Gu, Didier Schwab, and Laurent
  Besacier. 2020.
\newblock Dual-decoder transformer for joint automatic speech recognition and
  multilingual speech translation.
\newblock In \emph{Proceedings of the 28th International Conference on
  Computational Linguistics}, pages 3520--3533.

\bibitem[{Liu et~al.(2019)Liu, Xiong, Zhang, He, Wu, Wang, and
  Zong}]{liu19d_interspeech}
Yuchen Liu, Hao Xiong, Jiajun Zhang, Zhongjun He, Hua Wu, Haifeng Wang, and
  Chengqing Zong. 2019.
\newblock \href {https://doi.org/10.21437/Interspeech.2019-2582} {{End-to-End
  Speech Translation with Knowledge Distillation}}.
\newblock In \emph{Proc. Interspeech 2019}, pages 1128--1132.

\bibitem[{McAuliffe et~al.(2017)McAuliffe, Socolof, Mihuc, Wagner, and
  Sonderegger}]{mcauliffe2017montreal}
Michael McAuliffe, Michaela Socolof, Sarah Mihuc, Michael Wagner, and Morgan
  Sonderegger. 2017.
\newblock Montreal forced aligner: Trainable text-speech alignment using kaldi.
\newblock \emph{Proc. Interspeech 2017}, pages 498--502.

\bibitem[{Ott et~al.(2019)Ott, Edunov, Baevski, Fan, Gross, Ng, Grangier, and
  Auli}]{ott2019fairseq}
Myle Ott, Sergey Edunov, Alexei Baevski, Angela Fan, Sam Gross, Nathan Ng,
  David Grangier, and Michael Auli. 2019.
\newblock fairseq: A fast, extensible toolkit for sequence modeling.
\newblock In \emph{Proceedings of the 2019 Conference of the North American
  Chapter of the Association for Computational Linguistics (Demonstrations)},
  pages 48--53.

\bibitem[{Panayotov et~al.(2015)Panayotov, Chen, Povey, and
  Khudanpur}]{panayotov2015librispeech}
Vassil Panayotov, Guoguo Chen, Daniel Povey, and Sanjeev Khudanpur. 2015.
\newblock Librispeech: an asr corpus based on public domain audio books.
\newblock In \emph{2015 IEEE international conference on acoustics, speech and
  signal processing (ICASSP)}, pages 5206--5210. IEEE.

\bibitem[{Peyr{\'e} et~al.(2019)Peyr{\'e}, Cuturi
  et~al.}]{peyre2019computational}
Gabriel Peyr{\'e}, Marco Cuturi, et~al. 2019.
\newblock Computational optimal transport: With applications to data science.
\newblock \emph{Foundations and Trends{\textregistered} in Machine Learning},
  11(5-6):355--607.

\bibitem[{Pino et~al.(2020)Pino, Xu, Ma, Dousti, and Tang}]{pino20_interspeech}
Juan Pino, Qiantong Xu, Xutai Ma, Mohammad~Javad Dousti, and Yun Tang. 2020.
\newblock \href {https://doi.org/10.21437/Interspeech.2020-2938}
  {{Self-Training for End-to-End Speech Translation}}.
\newblock In \emph{Proc. Interspeech 2020}, pages 1476--1480.

\bibitem[{Post(2018)}]{post2018call}
Matt Post. 2018.
\newblock A call for clarity in reporting bleu scores.
\newblock In \emph{Proceedings of the Third Conference on Machine Translation:
  Research Papers}, pages 186--191.

\bibitem[{Salesky et~al.(2019)Salesky, Sperber, and Waibel}]{salesky2019fluent}
Elizabeth Salesky, Matthias Sperber, and Alexander~H Waibel. 2019.
\newblock Fluent translations from disfluent speech in end-to-end speech
  translation.
\newblock In \emph{NAACL-HLT (1)}.

\bibitem[{Schakel and Wilson(2015)}]{DBLP:journals/corr/SchakelW15}
Adriaan M.~J. Schakel and Benjamin~J. Wilson. 2015.
\newblock \href {http://arxiv.org/abs/1508.02297} {Measuring word significance
  using distributed representations of words}.
\newblock \emph{CoRR}, abs/1508.02297.

\bibitem[{So et~al.(2022)So, Oh, Lim, Byun, Shin, and Song}]{so2022geodesic}
Junhyuk So, Changdae Oh, Yongtaek Lim, Hoyoon Byun, Minchul Shin, and Kyungwoo
  Song. 2022.
\newblock Geodesic multi-modal mixup for robust fine-tuning.
\newblock \emph{arXiv preprint arXiv:2203.03897}.

\bibitem[{STENTIFORD and STEER(1988)}]{stentiford1988machine}
FWM STENTIFORD and MG~STEER. 1988.
\newblock Machine translation of speech.
\newblock \emph{British Telecom technology journal}, 6(2):116--123.

\bibitem[{Stoian et~al.(2020)Stoian, Bansal, and
  Goldwater}]{stoian2020analyzing}
Mihaela~C Stoian, Sameer Bansal, and Sharon Goldwater. 2020.
\newblock Analyzing asr pretraining for low-resource speech-to-text
  translation.
\newblock In \emph{ICASSP 2020-2020 IEEE International Conference on Acoustics,
  Speech and Signal Processing (ICASSP)}, pages 7909--7913. IEEE.

\bibitem[{Tang et~al.(2021)Tang, Pino, Wang, Ma, and Genzel}]{tang2021general}
Yun Tang, Juan Pino, Changhan Wang, Xutai Ma, and Dmitriy Genzel. 2021.
\newblock \href {https://doi.org/10.1109/ICASSP39728.2021.9415058} {A general
  multi-task learning framework to leverage text data for speech to text
  tasks}.
\newblock In \emph{ICASSP 2021 - 2021 IEEE International Conference on
  Acoustics, Speech and Signal Processing (ICASSP)}, pages 6209--6213.

\bibitem[{Vaswani et~al.(2017)Vaswani, Shazeer, Parmar, Uszkoreit, Jones,
  Gomez, Kaiser, and Polosukhin}]{vaswani2017attention}
Ashish Vaswani, Noam Shazeer, Niki Parmar, Jakob Uszkoreit, Llion Jones,
  Aidan~N Gomez, {\L}ukasz Kaiser, and Illia Polosukhin. 2017.
\newblock Attention is all you need.
\newblock \emph{Advances in neural information processing systems}, 30.

\bibitem[{Verma et~al.(2019)Verma, Lamb, Beckham, Najafi, Mitliagkas,
  Lopez-Paz, and Bengio}]{verma2019manifold}
Vikas Verma, Alex Lamb, Christopher Beckham, Amir Najafi, Ioannis Mitliagkas,
  David Lopez-Paz, and Yoshua Bengio. 2019.
\newblock Manifold mixup: Better representations by interpolating hidden
  states.
\newblock In \emph{International Conference on Machine Learning}, pages
  6438--6447. PMLR.

\bibitem[{Vila et~al.(2018)Vila, Escolano, Fonollosa, and
  Costa-Jussa}]{vila2018end}
Laura~Cross Vila, Carlos Escolano, Jos{\'e}~AR Fonollosa, and Marta~R
  Costa-Jussa. 2018.
\newblock End-to-end speech translation with the transformer.
\newblock In \emph{IberSPEECH}, pages 60--63.

\bibitem[{Villani(2009)}]{villani2009optimal}
C{\'e}dric Villani. 2009.
\newblock \emph{Optimal transport: old and new}, volume 338.
\newblock Springer.

\bibitem[{Vydana et~al.(2021)Vydana, Karafiát, Zmolikova, Burget, and
  Černocký}]{vydana2021jointly}
Hari~Krishna Vydana, Martin Karafiát, Katerina Zmolikova, Lukáš Burget, and
  Honza Černocký. 2021.
\newblock \href {https://doi.org/10.1109/ICASSP39728.2021.9414159} {Jointly
  trained transformers models for spoken language translation}.
\newblock In \emph{ICASSP 2021 - 2021 IEEE International Conference on
  Acoustics, Speech and Signal Processing (ICASSP)}, pages 7513--7517.

\bibitem[{Waibel et~al.(1991)Waibel, Jain, McNair, Saito, Hauptmann, and
  Tebelskis}]{waibel1991janus}
A~Waibel, AN~Jain, AE~McNair, H~Saito, AG~Hauptmann, and J~Tebelskis. 1991.
\newblock Janus: a speech-to-speech translation system using connectionist and
  symbolic processing strategies.
\newblock In \emph{[Proceedings] ICASSP 91: 1991 International Conference on
  Acoustics, Speech, and Signal Processing}, pages 793--796. IEEE.

\bibitem[{Wang et~al.(2020{\natexlab{a}})Wang, Tang, Ma, Wu, Okhonko, and
  Pino}]{wang2020fairseq}
Changhan Wang, Yun Tang, Xutai Ma, Anne Wu, Dmytro Okhonko, and Juan Pino.
  2020{\natexlab{a}}.
\newblock Fairseq s2t: Fast speech-to-text modeling with fairseq.
\newblock In \emph{Proceedings of the 1st Conference of the Asia-Pacific
  Chapter of the Association for Computational Linguistics and the 10th
  International Joint Conference on Natural Language Processing: System
  Demonstrations}, pages 33--39.

\bibitem[{Wang et~al.(2021)Wang, Wu, Pino, Baevski, Auli, and
  Conneau}]{wang21r_interspeech}
Changhan Wang, Anne Wu, Juan Pino, Alexei Baevski, Michael Auli, and Alexis
  Conneau. 2021.
\newblock \href {https://doi.org/10.21437/Interspeech.2021-1912} {{Large-Scale
  Self- and Semi-Supervised Learning for Speech Translation}}.
\newblock In \emph{Proc. Interspeech 2021}, pages 2242--2246.

\bibitem[{Wang et~al.(2020{\natexlab{b}})Wang, Wu, Liu, Yang, and
  Zhou}]{wang2020bridging}
Chengyi Wang, Yu~Wu, Shujie Liu, Zhenglu Yang, and Ming Zhou.
  2020{\natexlab{b}}.
\newblock Bridging the gap between pre-training and fine-tuning for end-to-end
  speech translation.
\newblock In \emph{Proceedings of the AAAI Conference on Artificial
  Intelligence}, volume~34, pages 9161--9168.

\bibitem[{Wang et~al.(2020{\natexlab{c}})Wang, Wu, Liu, Zhou, and
  Yang}]{wang2020curriculum}
Chengyi Wang, Yu~Wu, Shujie Liu, Ming Zhou, and Zhenglu Yang.
  2020{\natexlab{c}}.
\newblock Curriculum pre-training for end-to-end speech translation.
\newblock In \emph{Proceedings of the 58th Annual Meeting of the Association
  for Computational Linguistics}, pages 3728--3738.

\bibitem[{Xie et~al.(2020)Xie, Wang, Wang, and Zha}]{xie2020fast}
Yujia Xie, Xiangfeng Wang, Ruijia Wang, and Hongyuan Zha. 2020.
\newblock A fast proximal point method for computing exact wasserstein
  distance.
\newblock In \emph{Uncertainty in artificial intelligence}, pages 433--453.
  PMLR.

\bibitem[{Xu et~al.(2021)Xu, Hu, Li, Zhang, Huang, Ju, Xiao, and
  Zhu}]{xu2021stacked}
Chen Xu, Bojie Hu, Yanyang Li, Yuhao Zhang, Shen Huang, Qi~Ju, Tong Xiao, and
  Jingbo Zhu. 2021.
\newblock Stacked acoustic-and-textual encoding: Integrating the pre-trained
  models into speech translation encoders.
\newblock In \emph{Proceedings of the 59th Annual Meeting of the Association
  for Computational Linguistics and the 11th International Joint Conference on
  Natural Language Processing (Volume 1: Long Papers)}, pages 2619--2630.

\bibitem[{Yang et~al.(2021)Yang, Chen, Zhou, and Li}]{yang2021enhancing}
Huiyun Yang, Huadong Chen, Hao Zhou, and Lei Li. 2021.
\newblock Enhancing cross-lingual transfer by manifold mixup.
\newblock In \emph{International Conference on Learning Representations}.

\bibitem[{Ye and Guo(2022)}]{ye2022dual}
Junjie Ye and Junjun Guo. 2022.
\newblock Dual-level interactive multimodal-mixup encoder for multi-modal
  neural machine translation.
\newblock \emph{Applied Intelligence}, pages 1--10.

\bibitem[{Ye et~al.(2021)Ye, Wang, and Li}]{ye21_interspeech}
Rong Ye, Mingxuan Wang, and Lei Li. 2021.
\newblock \href {https://doi.org/10.21437/Interspeech.2021-1065} {{End-to-End
  Speech Translation via Cross-Modal Progressive Training}}.
\newblock In \emph{Proc. Interspeech 2021}, pages 2267--2271.

\bibitem[{Ye et~al.(2022)Ye, Wang, and Li}]{ye-etal-2022-cross}
Rong Ye, Mingxuan Wang, and Lei Li. 2022.
\newblock \href {https://doi.org/10.18653/v1/2022.naacl-main.376} {Cross-modal
  contrastive learning for speech translation}.
\newblock In \emph{Proceedings of the 2022 Conference of the North American
  Chapter of the Association for Computational Linguistics: Human Language
  Technologies}, pages 5099--5113, Seattle, United States. Association for
  Computational Linguistics.

\bibitem[{Yokoi et~al.(2020)Yokoi, Takahashi, Akama, Suzuki, and
  Inui}]{yokoi2020word}
Sho Yokoi, Ryo Takahashi, Reina Akama, Jun Suzuki, and Kentaro Inui. 2020.
\newblock Word rotator’s distance.
\newblock In \emph{Proceedings of the 2020 Conference on Empirical Methods in
  Natural Language Processing (EMNLP)}, pages 2944--2960.

\bibitem[{Zhang et~al.(2020)Zhang, Williams, Titov, and
  Sennrich}]{zhang2020improving}
Biao Zhang, Philip Williams, Ivan Titov, and Rico Sennrich. 2020.
\newblock Improving massively multilingual neural machine translation and
  zero-shot translation.
\newblock In \emph{Proceedings of the 58th Annual Meeting of the Association
  for Computational Linguistics}, pages 1628--1639.

\bibitem[{Zhang et~al.(2018)Zhang, Cisse, Dauphin, and
  Lopez-Paz}]{zhang2018mixup}
Hongyi Zhang, Moustapha Cisse, Yann~N Dauphin, and David Lopez-Paz. 2018.
\newblock mixup: Beyond empirical risk minimization.
\newblock In \emph{International Conference on Learning Representations}.

\bibitem[{Zhang et~al.(2022{\natexlab{a}})Zhang, Xu, Hu, Zhang, Xiao, and
  Zhu}]{zhang2022improving}
Yuhao Zhang, Chen Xu, Bojie Hu, Chunliang Zhang, Tong Xiao, and Jingbo Zhu.
  2022{\natexlab{a}}.
\newblock Improving end-to-end speech translation by leveraging auxiliary
  speech and text data.
\newblock \emph{arXiv preprint arXiv:2212.01778}.

\bibitem[{Zhang et~al.(2022{\natexlab{b}})Zhang, Zhou, Ao, Liu, Dai, Li, and
  Wei}]{zhang2022speechut}
Ziqiang Zhang, Long Zhou, Junyi Ao, Shujie Liu, Lirong Dai, Jinyu Li, and Furu
  Wei. 2022{\natexlab{b}}.
\newblock \href {https://doi.org/10.48550/ARXIV.2210.03730} {Speechut: Bridging
  speech and text with hidden-unit for encoder-decoder based speech-text
  pre-training}.

\bibitem[{Zhao et~al.(2021{\natexlab{a}})Zhao, Wang, Dong, Ye, and
  Li}]{zhao2021neurst}
Chengqi Zhao, Mingxuan Wang, Qianqian Dong, Rong Ye, and Lei Li.
  2021{\natexlab{a}}.
\newblock Neurst: Neural speech translation toolkit.
\newblock In \emph{Proceedings of the 59th Annual Meeting of the Association
  for Computational Linguistics and the 11th International Joint Conference on
  Natural Language Processing: System Demonstrations}, pages 55--62.

\bibitem[{Zhao et~al.(2021{\natexlab{b}})Zhao, Luo, Chen, and
  Gilman}]{zhao2021mutual}
Jiawei Zhao, Wei Luo, Boxing Chen, and Andrew Gilman. 2021{\natexlab{b}}.
\newblock Mutual-learning improves end-to-end speech translation.
\newblock In \emph{Proceedings of the 2021 Conference on Empirical Methods in
  Natural Language Processing}, pages 3989--3994.

\bibitem[{Zheng et~al.(2021)Zheng, Chen, Ma, and Huang}]{zheng2021fused}
Renjie Zheng, Junkun Chen, Mingbo Ma, and Liang Huang. 2021.
\newblock Fused acoustic and text encoding for multimodal bilingual pretraining
  and speech translation.
\newblock In \emph{International Conference on Machine Learning}, pages
  12736--12746. PMLR.

\end{thebibliography}
\bibliographystyle{acl_natbib}

\appendix

\end{document}